\documentclass{article}

\usepackage[nonatbib, final]{nips_2018}
\usepackage{bm}
\usepackage{graphicx}
\usepackage{floatrow}

\usepackage[utf8]{inputenc} 
\usepackage[T1]{fontenc}    
\usepackage{hyperref}       
\usepackage{url}            
\usepackage{booktabs}       
\usepackage{amsfonts}       
\usepackage{nicefrac}       
\usepackage{microtype}      
\usepackage{amsmath}
\usepackage{authblk}

\title{Semi-supervised Rare Disease Detection Using Generative Adversarial Network}

\author[1,2]{Wenyuan Li \thanks{Email: liwenyuan.zju@gmail.com}}
\author[1]{Yunlong Wang}
\author[1]{Yong Cai}
\author[3]{Corey Arnold}
\author[1]{Emily Zhao}
\author[1]{Yilian Yuan}
\affil[1]{Department of Advanced Analytic, IQVIA}
\affil[2]{Department of Electrical and Computer Engineering, University of California, Los Angeles}
\affil[3]{Departments of Radiological Sciences and Pathology and Laboratory Medicine, University of California, Los Angeles}

\begin{document}

\maketitle

\begin{abstract}
 Rare diseases affect a relatively small number of people, which limits investment in research for treatments and cures. Developing an efficient method for rare disease detection is a crucial first step towards subsequent clinical research. In this paper, we present a semi-supervised learning framework for rare disease detection using generative adversarial networks. Our method takes advantage of the large amount of unlabeled data for disease detection and achieves the best results in terms of precision-recall score compared to baseline techniques.
 
\end{abstract}

\section{Introduction}
A rare disease has a very low prevalence rate that affects only a small percentage of the population \cite{boat2011rare}. Finding patients with undiagnosed rare diseases is like looking for a needle in a haystack. Therefore, it is essential to develop efficient methods to help with detection and diagnosis in order to raise disease awareness and to provide early disease intervention \cite{cameron2010evaluation}. On the other hand, interest in machine learning for healthcare has grown immensely during last several years \cite{obermeyer2016predicting, ghassemi2018opportunities, ching2018opportunities}. Several machine learning methods, such as Recurrent Network \cite{choi2016using, lipton2015learning}, FHIR-formatted representation \cite{rajkomar2018scalable}, etc. have been proposed to predict patient-level disease. For more comprehensive overview of machine learning application on healthcare, we refer readers to \cite{obermeyer2016predicting, ghassemi2018opportunities, ching2018opportunities}. 

 Recently, generative adversarial networks (GANs) \cite{goodfellow2014generative}, were demonstrated in a semi-supervised learning (SSL) framework \cite{salimans2016improved, dai2017good}, where typically a small amount of labeled data with a large amount of unlabeled data are present. A GAN sets up two competing neural network models: a discriminator D and a generator G. The task of D is to classify an input as either the output from the generator (“fake” data), or actual samples from the underlying data ``manifold'' (``real'' data) and the class in which the samples belong. The goal of G is to produce outputs that are classified by the discriminator as ``real'', namely as coming from the underlying data ``manifold''. By taking advantage of the unlabeled data to estimate the shape of the data ``manifold'' and generating compliment ``fake'' data to force the decision boundary to be outside of the data ``manifold'', GANs demonstrate superior performance in SSL \cite{dai2017good, oliver2018realistic}. 
 
 The rare disease detection problem is a good fit for SSL, where we usually have limited labeled data that are available and aim to leverage the large amount of unlabeled data to find potential undiagnosed patients. In this work, we follow the spirit of \cite{salimans2016improved,dai2017good}, and propose a novel framework for semi-supervised rare disease detection using a GAN. The key contributions of this paper are: 1) we applied the SSL method to the rare disease detection problem using a GAN and achieved better results than a baseline model; 2) we carefully designed a loss function that helped G to generate complement samples of the underlying data distribution; and 3) we did an extensive ablation study on how the SSL method could help with rare disease detection and how each penalty term in the loss function would affect model performance.

\section{Methods}
\label{gen_inst}
\subsection{Data and pre-processing}

We extract data from IQVIA longitudinal prescription (Rx) and medical claims (Dx) database, including hundreds of millions patients' clinical records. To limit scope, we focus on only one particular rare disease, an inherited blood disorder caused by a genetic defect. It is estimated to affect about 1 in 20,000 people according to Genetic Home Reference (GHR) from NIH.

For model development, we pulled the diagnoses, procedures, and prescriptions at transaction level from January 1, 2010 to July 31, 2017. After applying eligibility rules, we keep only the patients with complete Rx/Dx information for our study. We label a patient with the rare disease condition as ``positive patients'', and without the rare disease as ``negative patients''. From the extracted data, we positively identified 5,923 true rare disease patients with valid records such as gender, age and 265 disease-relevant symptoms that were pre-selected by physicians. To boost the positive signal and model development, we constructed a training dataset by matching each positive patient with three randomly selected negative patients. The final training data contains 5,923 positive patients, 17,769 negative patients, and 1,166,831 unlabeled patients. The testing dataset contains 1,771,227 patients with 23,246 positive patients. The positive ratio in the testing data is about 0.013. Note that unlike the general literature of the similar problem in computer vision \cite{salimans2016improved, dai2017good, oliver2018realistic}, the unlabeld data in our case doesn't have any actual labels.

For each of the complete set of symptoms across all patient records, we calculated the time difference between the first record date and the last record date of that symptom for each patient. Note that if a patient is positive, only records before diagnosis are utilized to avoid information leakage. For patients who did not have the symptom, we used the value from the patient with the maximum time difference who had the symptom. The count and frequency for each symptom were also added as input features. Zeros were used to fill missing values for count and frequency as these patients never experienced certain symptoms. Time difference, count, and frequency for the 265 symptoms along with patient age and patient gender provided us with 797 features in total for each patient. Finally, for better training performances with GANs architecture, we linearly normalized each feature between -1 and 1, which removed potentially erroneous values and extreme outliers.   

\subsection{GAN model architecture}
We are given a labeled dataset, $\mathcal{L} = \left \{(\bm{x}, y)\right \}$, where $y \in \left \{0, 1\right \}$ denotes if the patient $\bm{x}$ has the rare disease ($y = 1$) or not ($y = 0$), and an unlabeled dataset $\mathcal{U}$ that only has the patients' features $\bm{x}$ but without label $y$. For rare disease detection, we want a simple classification model (\textit{e.g.} the discriminator in GAN in our case) to take in $\bm{x}$ as input and output a two-dimensional vector of logits $\left \{ l_0, l_1 \right \}$, that can be turned into class probabilities by softmax: $p_{D}(y = 1|\bm{x}) = \frac{exp(l_1)}{exp(l_0) + exp(l_1)}$ indicates the probability that the patient has the rare disease, and $p_{D}(y = 0|\bm{x}) = 1 - p_{model}(y = 1|\bm{x})$ indicates that the patient does not. For this, we follow \cite{salimans2016improved,dai2017good} to add samples from the unlabeled dataset $\mathcal{U}$ and the GAN's generator to our dataset. The unlabeled data are treated as ``real data'', while the generated data are labeled as a third ``generated'' class. We use $l_2$ and $p_{D}(y = 2|\bm{x})$ to denote the output logit and probability of the ``generated'' class for an input feature $\bm{x}$. Note that by subtracting a general function $f(\bm{x})$ from the output logits, the softmax probability does not change. Thus we can still formulate our discriminator in a binary fashion.

In our experiments, both the generator (G) and the discriminator (D) have five hidden layers. We use weight normalization \cite{salimans2016weight} and spatial drop out \cite{srivastava2014dropout} in both G and D. We add Gaussian noise \cite{sonderby2016amortised} to the output of each layer of D. A Tanh layer is added to the end of G that maps the output to the range between -1 and 1.

\subsection{Objective functions}\label{loss functions}
The objective of our model is to detect the presence of the rare disease given a patient's input features $\bm{x}$. For the discriminator, the loss consists of three different parts as discussed in \cite{salimans2016improved}:$L_{D} = L_{\mathcal{L}} + L_{\mathcal{U}} + L_{G}$, where 

\begin{equation}
\begin{aligned}
L_{\mathcal{L}} &= -\mathop{\mathbb{E}}_{\bm{x},y\sim\mathcal{L}} \log p_{D}(y|\bm{x}, y \leq 1)\\
L_{\mathcal{U}} &= -\mathop{\mathbb{E}}_{\bm{x}\sim\mathcal{U}} \log p_{D}(y\leq 1|\bm{x})\\
L_{G} &= -\mathop{\mathbb{E}}_{\bm{x}\sim G} \log p_{D}(y = 2|\bm{x})
\end{aligned}
\end{equation}

$L_{\mathcal{L}}$ is the ordinary cross-entropy loss that tries to minimize misclassification among the labeled data; $L_{\mathcal{U}}$ tries to maximize the log probability that indicates the unlabeled data is real data apart from those generated by the generator; and $L_{G}$ tries to maximize the log probability to map the generated data to the ``generated'' class. Note that we can fix the logit $l_2(\bm{x}) = 0 ~ \forall \bm{x}$ for the generated class and the output of the softmax remains the same. We also add an entropy loss for the discriminator as suggested by \cite{dai2017good}, 
$L_{ent} = \mathop{\mathbb{E}}_{\bm{x}\sim\mathcal{U}}\sum_{i = 0}^1p_{D}(i|\bm{x})\log p_{D}(i|\bm{x})$. However, empirically we found that it hurts the performance, which will be discussed in Section \ref{discussion}. 






For the generator, we use a feature map loss $L_{fm} = \left \| \mathop{\mathbb{E}}_{\bm{x}\sim\mathcal{G}}f(\bm{x}) - \mathop{\mathbb{E}}_{\bm{x}\sim\mathcal{U}}f(\bm{x}) \right \|^2$. As suggested by \cite{salimans2016improved} and \cite{dai2017good}, feature matching loss can help generate samples outside of the true data manifold due to its weak power in distribution matching. This property helps with the semi-supervised problem. Additionally, we also add a pull-away term

\begin{equation}\label{eq4}
\begin{aligned}
L_{pt} &= \mathop{\mathbb{E}}_{\bm{x} \sim G} [\sum_{j\neq i}(\frac{f(\bm{x_i})^T f(\bm{x_j})}{\left \| \bm{x_i} \right \| \left \| \bm{x_j} \right \|})^2] + \mathop{\mathbb{E}}_{\bm{x_i} \sim G, \bm{x_j} \sim \mathcal{L}} [\sum_{j,i}(\frac{f(\bm{x_i})^T f(\bm{x_j})}{\left \| \bm{x_i} \right \| \left \| \bm{x_j} \right \|})^2].
\end{aligned}
\end{equation}

The first term on the right hand side inspired by \cite{dai2017good} tries to orthogonalize the features in each mini-batch by minimizing the squared cosine similarity. Hence, it can help to increase the diversity of generated samples and avoid mode collapse. The second term tires to pull away the generated data distribution from the true data distribution, thus the generator can generate more samples outside of the labeled data manifold. As suggested by \cite{dai2017good}, samples from the low-density feature space could help with semi-supervised learning in GANs. Thus our generator loss finally becomes $L_{G} = L_{fm} + L_{pt}$.

\subsection{Evaluation of GAN outputs}
As the testing data are highly imbalanced (only 23,246 out of 1,771,227 samples are positive), prediction accuracy and ROC curve (receiver operating characteristic curve) fail to provide good measurements for the model performance. We thus use precision-recall curves and the area under the curve (PR-AUC) as an evaluation of model performance. The PR-AUC score is measured by a Riemann summation method that applies the trapezoidal rule. The detailed explanation of this method can be found in \cite{wiki:AUC}.

\section{Results}
\subsection{Main Results}
\begin{figure}
\floatbox[{\capbeside\thisfloatsetup{capbesideposition={right,center},capbesidewidth= 1.5in}}]{figure}[\FBwidth]
{\caption{PR and AUC scores for the proposed GAN model and other baseline models. Our proposed GAN model achieves the best result, $34.18\%$ in PR-AUC. While logistic regression (LR), neural network (NN(D)), random forest (RF), achieve $29.04\%$, $28.95\%$, and $10.51\%$ PR-AUC, respectively.}  \label{Figrue1}}
{\includegraphics[width=3.5in]{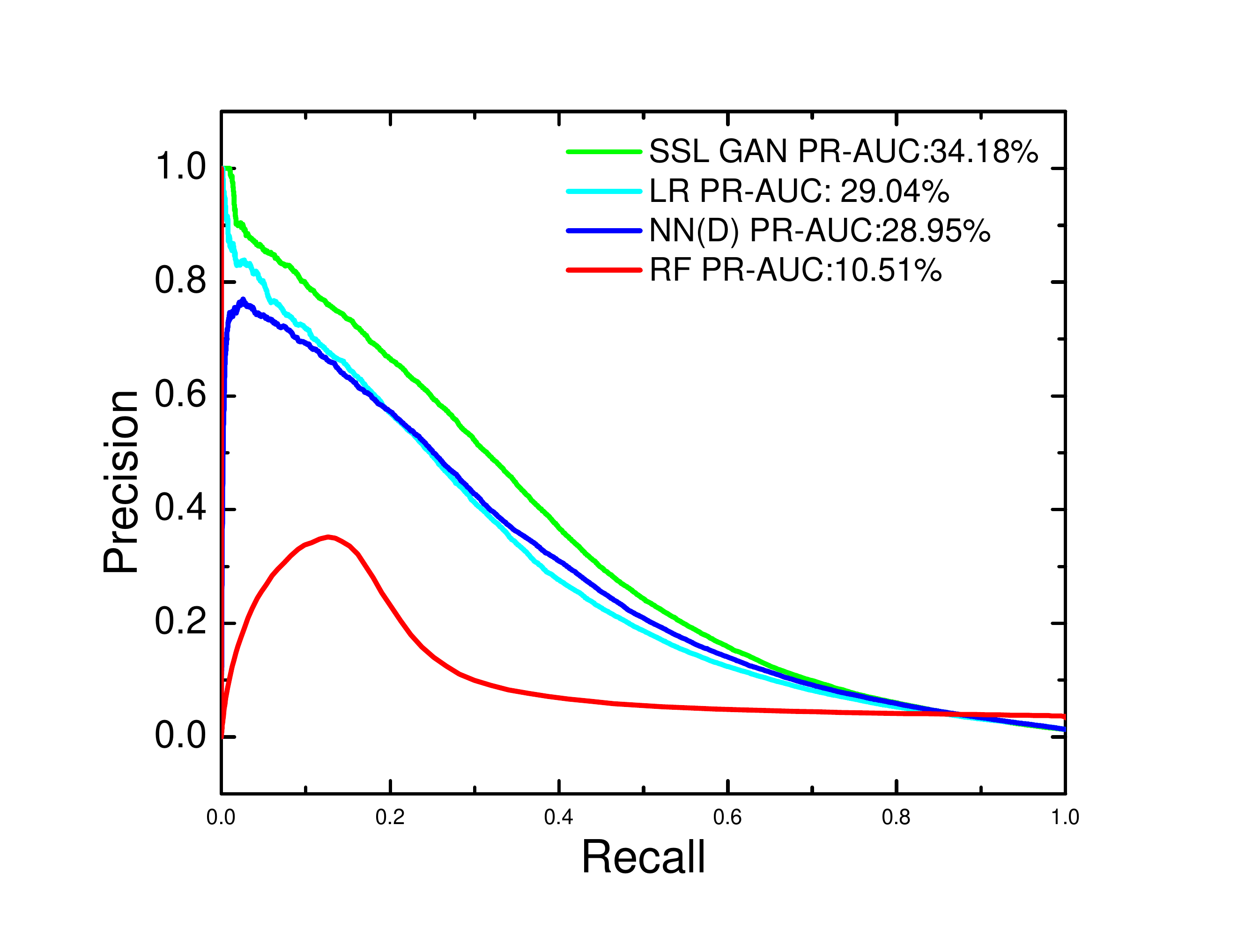}}
\end{figure}

We compare the results of our best model with three different classification methods: logistic regression (LR), neural network (NN(D)\footnote{NN(D) denotes that the neural network has the same architecture as in our discriminator D.}), and random forest (RF). Our proposed GAN model achieved the best result, $34.18\%$, in terms of the PR-AUC score. LR and NN(D) achieves relatively similar results, around $29\%$ in PR-AUC. While RF gives the poorest result, of only $10.51\%$ PR-AUC. The PR curves are shown in Fig.\ref{Figrue1}.

\subsection{Ablation Study}
We report the results of the ablation study in Table \ref{Table1}. NN(D) can be understood as the discriminator alone in the GAN model. The original GAN model is the GAN model without SSL branch \cite{goodfellow2014generative}, while SSL GAN stands for the semi-supervised GAN \cite{salimans2016improved,dai2017good}. FM is the feature matching. PT is the pull away term in Eq.\ref{eq4}. Ent refers to the conditional entropy term introduced by \cite{dai2017good}. Further discussion on the effects of components in our model can be found in Section \ref{discussion}. Note that the ablation study may sometimes be sensitive to the labeled/unlabeled split. However, we don't have any labeled information about the unlabeled data in the current scenario, therefore we will leave the related study in the future work.

\begin{table}[ht]
\caption{Ablation study.}
\label{Table1}
\centering
\begin{tabular}{lc|lc}
\hline
\multicolumn{1}{c}{Setting} & PR AUC Score & \multicolumn{1}{c}{Setting} & PR AUC Score     \\ \hline
NN(D)          & 28.95\%      & SSL GAN FM + PT             & \textbf{34.18\%} \\
Original GAN                & 29.08\%      & SSL GAN FM + Ent            & 30.20\%          \\
SSL GAN FM                  & 32.06\%      & SSL GAN FM + PT + Ent       & 30.33\%          \\ \hline
\end{tabular}
\end{table}

\section{Discussion}\label{discussion}
The results in Fig.\ref{Figrue1} show the power of unlabeled data. Our proposed model provides a way to leverage the large amount of unlabeled data, improving performance by roughly $5\%$ in terms of PR-AUC. The competing methods can only take advantage of the labeled data, which results in a worse performance. Note that in the evaluation, we are identifying 23,246 positive samples out of 1,771,227 (ratio $= 1.3\%$) for a PR-AUC of $34.18\%$, which is a significant improvement given the severe imbalance of the data.

The results of the ablation study in Table \ref{Table1} provide us with two insights. First, the large amount of unlabeled data is again key for good performance. Models without unlabeled data (NN(D) and Original GAN) achieve relatively low PR-AUC scores. Second, designing a good loss function is essential to achieve the best result. FM plus PT as discussed in Section \ref{loss functions} forms the best design in our case. FM helps the generator generate samples close to the data manifold, and PT helps the generator generate more samples from the low-density boundary and increase the diversity of the generated samples. We also add the Ent term in our model. However, empirically we found that it hurts model performance. A possible explanation by \cite{dai2017good} relates to the classic exploitation-exploration trade-off. Further theoretical explanations and experiments, beyond the scope of this paper, need to be done to provide more insights about the Ent term.

\section{Conclusion}
In this work, we present a semi-supervised learning framework for rare disease detection using generative adversarial network. We design a loss function that helps generate samples that reside in the complement of the real data ``manifold'', and achieves 34.18\% in PR-AUC, around 5\% higher compared to our baseline model. Extensive ablation study shows that generated data and unlabeled data help with the model performance by forcing the logits of the true classes to be low outside of the data ``manifold'' and providing the model with more hints on real data distribution. Our proposed method has potential for future uses in detecting rare diseases, as well as other areas, such as medical image analysis, in which scenario labeling data is time-consuming and expensive.



\bibliographystyle{IEEEtran}
\bibliography{IEEEabrv,Bibliography}
\end{document}